\documentclass[preprints,article,accept,moreauthors,pdftex]{Definitions/mdpi} 

\usepackage{algorithm,algpseudocode}

\firstpage{1} 
\pubvolume{xx}
\issuenum{1}
\articlenumber{5}
\pubyear{2019}
\copyrightyear{2019}
\history{Received: date; Accepted: date; Published: date}
\graphicspath{ {./images/} }

\Title{Differentiable programming: Generalization, characterization and limitations of deep learning}

\Author{Adri\'{a}n Hern\'{a}ndez $^{1}$, Gilles Millerioux$^{2}$ and Jos\'{e} M. Amig\'{o} $^{1,\dagger}$}

\AuthorNames{Firstname Lastname, Firstname Lastname and Firstname Lastname}

\address{
$^{1}$ \quad Centro de Investigaci\'{o}n Operativa, Universidad Miguel Hern\'{a}ndez, Av. de la Universidad s/n, 03202 Elche, Spain\\
$^{2}$ \quad Universit\'{e} de Lorraine, CNRS, CRAN, 2 Rue Jean Lamour, 54519, Vandoeuvre-les-Nancy, France}
\firstnote{Corresponding author} 
\corres{Correspondence: jm.amigo@umh.es}

\abstract{In the past years, deep learning models have been successfully applied in several cognitive tasks. Originally inspired by neuroscience, these models are specific examples of differentiable programs. In this paper we define and motivate differentiable programming, as well as specify some program characteristics that allow us to incorporate the structure of the problem in a differentiable program. We analyze different types of differentiable programs, from more general to more specific, and evaluate, for a specific problem with a graph dataset, its structure and knowledge with several differentiable programs using those characteristics. Finally, we discuss some inherent limitations of deep learning and differentiable programs, which are key challenges in advancing artificial intelligence, and then analyze possible solutions}


\begin{document}

\section{Introduction}
\label{sec1}

In recent years, different deep learning models (neural networks, RNNs, CNNs,...) have been developed and successfully applied in cognitive tasks such as natural language processing, gaming, computer vision and more \cite{LeCun2015DeepLearning}. Although these models were originally biologically inspired, they are specific examples of differentiable programs and the key to their success are the relationships and transformations of the elements (tensors) of the model. A differentiable program stands here for any tensor transformation whose parameters (trained end-to-end) are differentiable.

Differentiable programming is a programming framework that is expressive enough to contain the space of all functions, and flexible enough to incorporate the priors and the structure of the problem when we define a new program. In traditional programming there are certain techniques for constructing algorithms (divide and conquer, recursion, dynamic programming,...) that depend on the problem to be solved. For differentiable programming, it is interesting to have also some characteristics that allow us to incorporate the structure and priors of the problem at hand.

To be more specific, by the structure and priors of a problem we mean the knowledge we have of the problem: temporal dependency, symmetries, relationship between its components, etc. The program characteristics are a series of properties and relations between the elements (tensors) of a differentiable program that allow us to incorporate this prior knowledge and reduce the hypothesis space in which learning is carried out.

In this paper we define and motivate differentiable programming, as well as specify some program characteristics that are important to construct differentiable programs and adapt them to important classes of problems. We analyze several domains or types of differentiable programs, from more general to more specific, and map the problem knowledge and structure to these programs using such characteristics. This procedure will be exemplified with the CiteSeer dataset \cite{yang2016revisiting} and the problem of classifying scientific publications into one of six classes.

We also discuss the limitations of differentiable models and possible solutions, which is a central theme for the future of artificial intelligence. We explain the need of learning strategies that leverage previous data and models to generate new experiences and knowledge as humans do, providing some characteristics and examples of these strategies.

In sum, the topic of this paper is differentiable programming, which is a programming model that generalizes deep learning. We call the concrete examples or models that it generates differentiable programs, which are represented by acyclic directed graphs. The learning process is done in two steps: In the forward pass, the graph is built; in the backward pass, the parameters are adjusted by gradient descent.

This paper is organized as follows. In the next section we describe, motivate and define differentiable programming. In Section \ref{sec3} we discuss its flexibility and propose a characterization. In Section \ref{sec4} we study general programs and programs that incorporate very specific priors, while in Section \ref{sec5} we perform a series of experiments to analyze a particular graph problem using the defined characteristics. Finally, in Section \ref{sec6} we describe the limitations of differentiable models and possible solutions. The conclusions and a list of acronyms winds up this paper.

\section{Differentiable programming. Motivation and definition}
\label{sec2}

In the past years we have seen major advances in the field of machine learning. Deep neural networks, along with the computational capabilities of Graphics Processing Units (GPUs) \cite{Yadan2013MultiGPU} have improved the performance of several tasks, including image recognition, machine translation, language modelling, time series prediction and game playing \cite{LeCun2015DeepLearning, Sutskever2014SequenceTS, Silver2017MasteringTG}. 

Let us recall that, in a feedforward neural network (FNN) composed of multiple layers, the output (without the bias term) at layer $l$ is given by

\begin{equation}  \label{eq:mnn}
\boldsymbol{x}^{l+1}=\sigma(W^{l}\boldsymbol{x}^{l}),
\end{equation}

$W^{l}$ being the weight matrix at layer $l$. $\sigma$ is the activation function and $\boldsymbol{x}^{l+1}$ is the output vector at layer $l$ and the input vector at layer $l+1$. The weight matrices for the different layers are the parameters of the model.

\textit{Deep learning} is a part of machine learning that is based on neural networks and uses multiple layers, where each layer extracts higher level features from the input.

Although deep learning can implicitly implement logical reasoning \cite{Hohenecker2018Onto,SATNet2019,AssessingSATNet2020}, it has limitations that make it difficult to achieve more general intelligence. Among such limitations, let us mention that deep learning only performs perception and does not carry out conscious and sequential reasoning \cite{Marcus2018Deep}.

Simplified, the learning problem can be formulated as follows. Given an input space $\mathcal{X}$ and an output space $\mathcal{Y}$, the pairs $(X,Y)\in \mathcal{X} \times \mathcal{Y}$ being random variables distributed according to a joint probability mass or density function $\rho(x,y)$, construct a function $g:\mathcal{X} \rightarrow \mathcal{Y}$ from a hypothesis space of functions $\mathcal{H}$ which predicts $Y$ from $X$ after observing a sequence of $n$ pairs $(x_i,y_i)$ independently and identically distributed according to $\rho(x,y)$.

A learning algorithm over $\mathcal{H}$ is a computable map from $(X,Y)$ to $\mathcal{H}$, specifically, an algorithm that takes as input the sequence of training samples and outputs a function $g:\mathcal{X} \rightarrow \mathcal{Y} \in \mathcal{H}$ with a low probability of error $P(g(X)\neq Y$.

As a general rule we are interested in having a hypothesis space (set of functions) $\mathcal{H}$ as expressive as possible. However, to solve a specific problem, it is necessary to restrict this hypothesis space depending on the structure and priors of the problem. 

This is what we have seen in recent years with the development of specific models and graph structures. For example, multilayer neural networks are based on the compositional and hierarchical structure of information, RNNs (Recurrent Neural Networks) on temporal dependence, CNNs (Convolutional Neural Networks) on translational invariance, etc.

Therefore, to advance and generalize deep learning, a programming model or framework with the following characteristics is necessary.
\begin{enumerate}
\item Be expressive enough to define the space of all hypothesis or functions.
\item Be able to incorporate the structure and the priors of the problem we want to model. As stated in \cite{Hernandez2021Chapter}, incorporating prior knowledge about the underlying task or class of functions can make the learning process more efficient.
\item Be able to define new primitives to advance deep learning capabilities (reasoning, attention, memory, modeling of physical problems, etc.)
\end{enumerate}

As we are going to see, a natural evolution to move forward in these directions is to generalize deep learning using a framework composed of differentiable blocks that allows to overcome the limitations of classical models.

This approach, called differentiable programming, adds new differentiable components to traditional neural networks. For the purposes of this article, differentiable programming is defined as follows.

\begin{Definition}[Differentiable programming] \label{Def1}
Differentiable programming is a programming model defined by a tuple $\left \langle  n,l,E,f_i,v_i,\alpha_i \right \rangle$ where:
\begin{enumerate}
\item Programs are directed acyclic graphs.

\item Graph nodes are mathematical functions or variables and the edges correspond to the flow of intermediate values between the nodes.

\item $n$ is the number of nodes and $l$ the number of input variables of the graph, with $1\leq l< n$. $v_i$ for $i \in \{1,...,n\}$ is the variable associated with node $i$.

\item $E$ is the set of edges in the graph. For each directed edge $(i,j) \in E$ from node i to node j we have $i<j$, therefore the graph is topologically ordered.

\item $f_i$ for $i \in \{(l+1),...,n\}$ is the differentiable function computed by node $i$ in the graph. $\alpha_i$ for $i \in \{(l+1),...,n\}$ contains all input values for node $i$.

\item The forward algorithm or pass, given input variables $v_1,...,v_l$ calculates $v_i=f_i(\alpha_i)$ for $i=\{(l+1),...,n\}$.

\item The graph is dynamically constructed and composed of functions that are differentiable and whose parameters are learned from data.

\end{enumerate}
\end{Definition}

To learn the parameters of the graph from the data, automatic differentiation is used. Automatic differentiation, in its reverse mode and in contrast to manual, symbolic and numerical differentiation, computes the derivatives in a two-step process \cite{Baydin2018AutomaticDiff, Wang2018DemystifyingDP}. 
As described in \cite{Baydin2018AutomaticDiff}, a function $f:R^{n}\rightarrow R^{m}$ is constructed with intermediate variables $v_{i}$ such that:
\begin{enumerate}
\item variables $v_{i-n}=w_{i}, i=1,...,n$ are the parameters.
\item variables $v_{i}, i=1,...,l$ are the intermediate variables.
\item variables $y_{m-i}=v_{l-i}, i=m-1,...,0$ are the output variables.
\end{enumerate}

In a first step, the graph is built populating intermediate variables $v_i$ and recording the dependencies. In a second step, called the backward pass, derivatives are calculated by propagating for the output $y_j$ being considered, the adjoints $\overline{v}_{i}=\frac{\partial {y}_j}{\partial {v}_{i}}$ from the output to the inputs.

The reverse mode is more efficient to evaluate for functions with a large number of inputs (parameters) and a small number of outputs. When $f:R^{n}\rightarrow R$, as is the case in machine learning with $n$ very large and $f$ the cost function, only one pass of the reverse mode is necessary to compute the gradient $\nabla f=(\frac{\partial y}{\partial w_1},...,\frac{\partial y}{\partial w_n}).$ 
 
In the last years, deep learning frameworks such as PyTorch have been developed that provide reverse-mode automatic differentiation \cite{Paszke2017automatic}. The define-by-run philosophy of PyTorch, whose
execution dynamically constructs the computational graph, facilitates the development of general differentiable programs.

Differentiable programming can be seen as a continuation of the deep learning end-to-end architectures that have replaced, for example, the traditional linguistic components in natural language processing \cite{Deng2018NLP,
Goldberg2017NLPBook}. Differentiable programs are composed of classical blocks (feedforward, recurrent neural networks, etc.) along with new ones such as differentiable branching, attention, memories, etc.

Differentiable programming is an evolution of classical (traditional) software programming where, as summarized in Table \ref{table1}:
\begin{enumerate}
\item Instead of specifying explicit instructions to the computer, an objective is set and an optimizable architecture is defined which allows to search in a subset of possible programs.
\item The program is defined by the input-output data and not predefined by the user.
\item The optimizable elements of the program have to be differentiable, say, by converting them into differentiable blocks. 
\end{enumerate}

\begin{table}[h!]
\begin{center}
    \begin{tabular}{l|l|l} % <-- Changed to S here.
      \textbf{Classical Programming} & \textbf{Differentiable Programming}\\
      \hline
      Sequence of explicit instructions & Sequence of differentiable primitives\\
      Fixed architecture & Optimizable architecture that searchs in a\\&subset of possible programs\\
      User defined programs & Data defined programs\\
      Imperative programming & Declarative programming, specifying the objectives\\& but not how to achieve them \\
      Direct, intuitive and explainable& High level of abstraction\\   
    \end{tabular}
\end{center}
\caption{Differentiable vs classical programming.}
\label{table1}
\end{table}

RNNs, for example, are an evolution of feedforward networks because they are classical neural networks inside a for-loop (a control flow statement for iteration) which allows the neural network to be executed repeatedly with recurrence. However, this for-loop is a predefined feature of the model. Differentiable programming allows to dynamically constructs the graph and vary the length of the loop. The ideal situation would be to augment the neural network with programming primitives (for-loops, if branches, while statements, external memories, logical modules, etc.) that are not predefined by the user but learned with the training data.

\begin{figure}[h]
\centering
\includegraphics[scale=0.95]{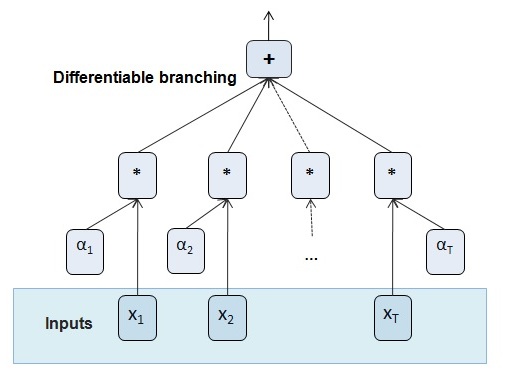} % Give a unique label
\caption{Differentiable program (computational graph) of differentiable branching.}
\label{DiffBranching}
\end{figure}

But many of these programming primitives are not differentiable and need to be converted into optimizable modules. For instance, if the condition $a$ of an "if" primitive (e.g., if $a$ is satisfied do $y(x)$, otherwise do $z(x)$) is to be learned, it can be the output of a neural network (linear transformation and a sigmoid function) and the conditional primitive will transform into a convex combination of both branches $ay(x)+(1-a)z(x)$. Similarly, in an attention module, different weights that are learned with the model are assigned to give a different influence to each part of the input. Figure \ref{DiffBranching} shows the program (directed acyclic graph) of a conditional branching.

%%%%%%%%%%%%%%%%%%%%%%%%%%%%%%%%%%%%%%%%%%

\section{Flexibility and characterization}
\label{sec3}
From Definition \ref{Def1}, it can be seen that differentiable programs are very flexible and give rise to very varied structures, which allows restricting the hypothesis space in an adequate way.

The following elements of differentiable programs support such flexibility and variability.
\begin{enumerate}
    \item Nodes can be tensors of any rank. E.g. scalars (0-rank tensors) ${v}$, vectors (1-rank) ${v}_{i}$, matrices (2-rank tensors) ${v}_{i}^{j}$, general tensors ${v}_{i_1i_2...}^{j_1j_2...}$. 
    \item The function $f_i$ computed by each node $i$ can be any transformation of the input tensors, as long as the program (composite of functions $f_i$)  can represent any continuous function (universal approximator). If the function $f_i$ has parameters, it has to be differentiable with respect to the parameters. 
    Examples are linear transformation of a vector $A\boldsymbol{v}$, a non-linear function of a vector $\sigma(A\boldsymbol{v})$ , a weighted sum of vectors $\sum\alpha_{i}\boldsymbol{v}_{i}$, etc. 
    \item The program not only performs a transformation of the input but also can integrate information in a very flexible way. As we will see in the next subsection, information nodes can be integrated and related using different graph paths.
    \item The graph is dynamically constructed for each input at execution. To do this, each of the primitives are executed sequentially, be they classic programming primitives (for, while, if, etc.) or differentiable functions.
    Therefore, the graph can be different depending on the input.
    \item Parameters can be set at training time or dynamically calculated. For example, the attention weights $\alpha_n$ in Figure \ref{DiffBranching} usually depend on both a fixed parameter and the input.
\end{enumerate}

Thus, differentiable programming generates, from a general hypothesis space $\mathcal{H}$ and a set of priors $\mathcal{P}_i$, a restricted hypothesis space $\mathcal{H}_1\subset \mathcal{H}$. The learning algorithm over $\mathcal{H}_1$ takes as input the sequence of training samples, builds the graph, computes the derivatives and outputs a program or function $g \in \mathcal{H}_1$.

Precisely, the flexibility and richness of differentiable programming make it easy to generate the restricted hypothesis space $\mathcal{H}_1$, translating the prior knowledge and structure of the problem to the elements of a differentiable program, i.e. to a tuple $\left \langle  n,l,E,f_i,v_i,\alpha_i \right \rangle$.

Next, we define some program characteristics that help in this process. They are also instrumental to define differentiable programs and adapt them to important classes of problems.

\subsection{Program characteristics}
\label{sec31}

In traditional programming there are certain heuristics for constructing algorithms depending on the problem to be solved. New structures have also been developed in neuroscience that help explain neural processing \cite{Hernandez2018MultilayerAN}. 

For differentiable programming, it is also useful
to have some characteristics that allow us to incorporate the structure and priors of the considered problem.

Traditionally, the characteristics of differentiable and deep learning programs (multilayer feed-forward networks, convolutional networks, etc.) were inspired by the brain. However, as we are going to see, although their source of inspiration was the brain, the key to their success has been the concrete relationships and transformations of the elements (tensors) of a differentiable program.

These characteristics are not independent but related to each other. Here we define some of these characteristics.
\begin{enumerate}
\item Integration and relation of information tensors (vectors). Given a sequence of input vectors $\boldsymbol{x}=(\boldsymbol{x}_1,...,\boldsymbol{x}_T)$ with $\boldsymbol{x}_t \in \mathbb{R}^n$  and a sequence of intermediate or output vectors $\boldsymbol{y}=(\boldsymbol{y}_1,...,\boldsymbol{y}_{T`})$ with $\boldsymbol{y}_t \in \mathbb{R}^m$, a program characteristic is the graph path relationship (shortest path, non-intersecting paths...) between each vector. 

For example, in RNNs the length of the path between the output vector $\boldsymbol{y}_t$ and the past input vector $\boldsymbol{x}_{t-l}$ increases with $l$.    

\item Relationship between the represented problem structure (locality, local or distant relations, node degree, temporal relations, etc.) and the differentiable program characteristics (depth, edges information, temporal evolution, etc.).

An example of this characteristic is the relationship between the average degree of a graph and the depth (number of layers) of its corresponding graph neural network (GNN).

\item Invariant transformations of data and symmetries. If $f$ represents the function that transforms the data $x$ and $T$ is a function that performs a transformation of the data, $f$ is invariant under $T$ if and only if $f(T(x)=f(x)$.

For example, convolutional networks (CNNs) have translational invariance, that is, the model produces the same response despite translations of input elements.

In this way, we restrict the structure of the differentiable program taking into account certain symmetries of the information. Another example would be a transformation of input vectors that is invariant under a change of order or permutation of the vectors. Given a sequence of input vectors $\boldsymbol{x}=(\boldsymbol{x}_1,...,\boldsymbol{x}_T)$, the transformation $f(\boldsymbol{x})$ does not depend on the ordering of the set of vectors $\boldsymbol{x}_1,\boldsymbol{x}_2,...,\boldsymbol{x}_T$.

\item Combination of modules or tasks, that is, the possibility of combining different modules using classical or differentiable primitives. A model $M$ is composed of different parts or tasks $T_i$ and each task $T_i$ has a direct meaning or some grounding.

Thus, by imposing restrictions on the structure, it is possible to better approximate the real distribution of the problem and be able to carry out subsequent tasks.

For example, different particles that interact with each other in a physical model could correspond to the different nodes of a graph neural network. Also, as in \cite{hudson2018compositional}, a task can be decomposed into a series of reasoning steps, learning to perform iterative reasoning.

\end{enumerate}

In Section \ref{sec4} we are going to see several domains or types of differentiable programs, in which we analyze and map the problem structure to specific programs using the characteristics defined above.

\section{From general to specific differentiable programs}
\label{sec4}
\subsection{Generalization of neural networks. Self-Attention}
\label{sec41}

In describing the flexibility of differentiable programming, we stated that the node function $f_i$ can be any transformation of the input tensors, as long as the program can represent any continuous function (universal approximator).

Traditionally, following the inspiration of biological neural networks, a linear transformation (together with the activation function) of a vector has been used, as described in Equation (\ref{eq:mnn}). 

\begin{figure}[h]
\centering
\includegraphics[scale=0.65]{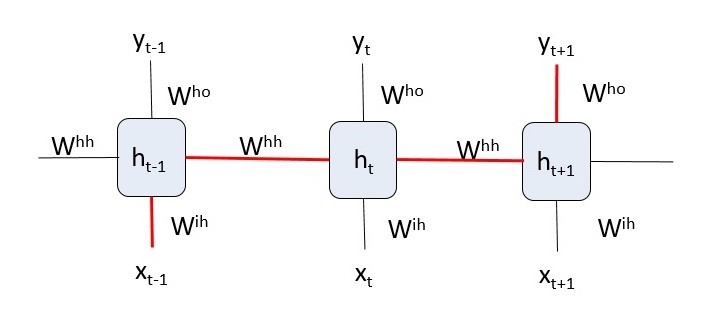} % Give a unique label
\caption{Computational graph of a recurrent neural network.}
\label{RNN}
\end{figure}

If we want to transform a set of vectors, then we can concatenate the vectors and use the neural network of Equation (\ref{eq:mnn}) (with a relevant increase in the number of parameters) or use a recurrent neural network (RNN). However, as seen in Figure \ref{RNN}, an RNN assumes a one-step temporal dependence, which implies that the length of the path between the output vector $\boldsymbol{y}_{t+1}$ and the past input vector $\boldsymbol{x}_{t-l}$ increases with $l$. Consequently, the relationship between the output $(\boldsymbol{y}_1,...,\boldsymbol{y}_T)$ and the input vectors $(\boldsymbol{x}_1,...,\boldsymbol{x}_T)$ is not symmetric. As a result, this model is biased towards certain problem structures.

\begin{figure}[h]
\centering
\includegraphics[scale=0.95]{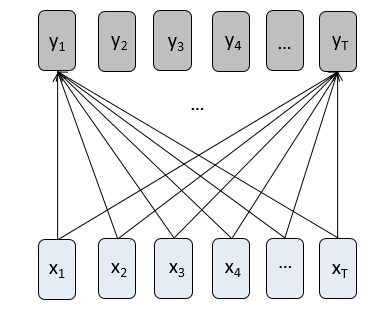} % Give a unique label
\caption{A direct and symmetric transformation of input vectors.}
\label{Direct}
\end{figure}

Therefore, a more general transformation of the input vectors should have the following characteristics.
\begin{enumerate}
\item It acts on a set of vectors, consciously and dynamically choosing the most important ones.
\item It has to integrate all the input vectors in a direct and symmetric way (as seen in Figure \ref{Direct}), with the same path length between an output vector and each of the input vectors.
\item The weights to integrate each input vector should not be fixed but dependent on the context (input).
\end{enumerate}

The structure described above is the basis of self-attention \cite{vaswani2017attention}, one of the most successful deep learning models developed in recent years. The input vectors are transformed (using learnable matrices) into query ($q$), key ($k$) and value ($v$) vectors and, for each step, the output is a direct integration of the value vectors based on the similarity between the query and each of the keys, $y_t=\sum_{i=1}^{T}similarity(\boldsymbol{q_t},\boldsymbol{k_i})\boldsymbol{v_i}$. Furthermore, this model has made it possible to integrate sequential and conscious reasoning into deep learning models \cite{HernandezAttention}.

\subsection{Compositional structures and reasoning}
\label{sec42}

The problems that we want to model using machine learning usually have a certain structure and are made up of a sequence of different tasks or modules. For example, human reasoning sometimes takes place in a series of phases, with each phase focusing on different information and feeding the next phase.

\begin{figure}[h]
\centering
\includegraphics[scale=0.75]{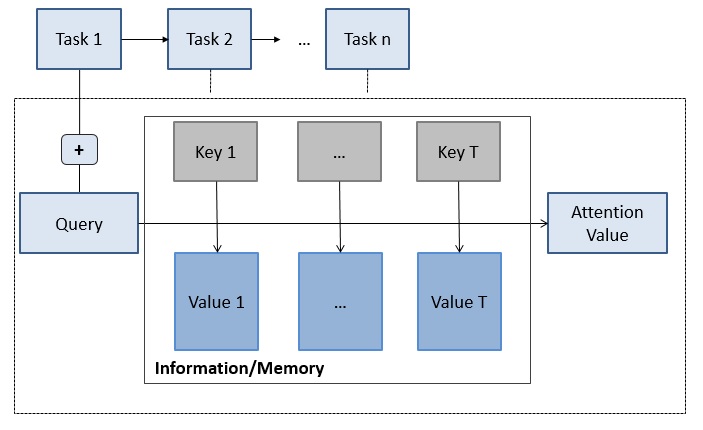} % Give a unique label
\caption{A sequence of attention tasks.}
\label{AttSequence}
\end{figure}

To make learning more efficient, the same structure can be transferred to the differentiable program, which will be made up of a sequence of modules. For example, a problem can be decomposed into a sequence of attention tasks. Each task focuses on a set of elements of information (text, image, memory, etc.) and is the input of the next task, as depicted in Figure \ref{AttSequence}.

Thus, a differentiable program can be very flexibly defined as a sequence or combination of tasks or modules. As stated in \cite{HernandezAttention}, an attention module can be added to any deep learning model in multiple ways: in several parts of the model; focusing on different elements; with different attention dimensions (spatial, temporal and input dimension), etc. 

Even more, as in Recurrent Independent Mechanisms (RIMs) \cite{Goyal2021RecurrentIM}, there can be multiple modules that operate independently and on each step compete with each other to read from the input and update their states, then attending only the parts of the input relevant to that module. 
\subsection{Graph structure restrictions. Graph neural networks}
\label{sec43}
After the success of deep learning in image processing, time series analysis, etc., it is logical to apply it to unstructured data and graphs. 

Graph neural networks \cite{PyGeometric2019,DynamicPyGeometric2021} apply neural networks to graphs, based on nodes connected to each other. To do this, each state $x$ of the node $v$ of the graph $G=(V,E)$ is iteratively updated by adding information from the neighboring nodes $\mathcal{N}(v)$:

\begin{equation}  \label{eq:Gnn}
\mathbf{x}_v^{k} = f^{k}_{\theta}(\mathbf{x}_v^{k-1},\Gamma_{j\in\mathcal{N}(v)}\phi^{k}(\mathbf{x}_v^{k-1},\mathbf{x}_j^{k-1},\mathbf{e}_{jv})) ,
\end{equation}

with $\mathbf{x}_v^{k-1}$ denoting F-dimensional vector features of node $v$ in layer $k-1$ and $\mathbf{e}_{jk}$ denoting optional edge features from node $j$ to node $v$. $\Gamma$ denotes a differentiable, permutation invariant function (e.g. a summation) and $f$ and $\phi$ denote differentiable functions such as feedforward neural networks (FNN). 

\begin{figure}[h]
\centering
\includegraphics[scale=0.85]{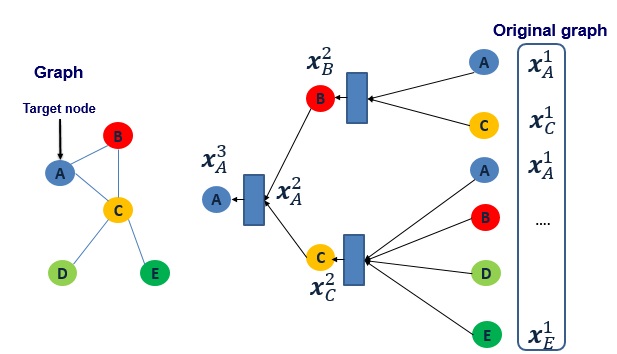} % Give a unique label
\caption{Two consecutive updates of information between nodes of a graph.}
\label{AggregationGNN}
\end{figure}

In Figure \ref{AggregationGNN}, where $\mathbf{x}_v^{1}$ represents the vector features of node $v$ in the original graph,  we can see two consecutive updates of information among the nodes of the graph.

Based on that definition, certain relationships can be drawn between the structure of the represented graph and the characteristics of GNNs:
\begin{enumerate}
\item Information aggregation in the represented network is performed via the layers in the GNN, that add
information from the neighboring nodes. The more layers, the further information travels from a node. 
\item Small world networks (most nodes are separated by a short distance) do not need many layers.
\item In theory, in graphs with a high average degree a GNN with few layers is enough to propagate the information of the nodes. However, the high average degree can cause saturation since the GNN nodes receive information from many nodes at the same time. 
\item Although increasing the number of layers of the GNN allows information to propagate between nodes, the task we perform at each node (e.g., prediction) may depend on the local neighborhood of a node or on distant information. Hence, two different graphs or problems but with similar structure (similar adjacency matrix) may require different GNNs.  
\item  There are some invariant transformations of information. For example, the function that integrates the information of the neighboring nodes (e.g. summation) should be permutation invariant.
\end{enumerate} 

%%%%%%%%%%%%%%%%%%%%%%%%%%%%%%%%%%%%%%%%%%

\section{Experiments}
\label{sec5}

\subsection{Materials and Methods}
\label{sec51}
In this section we evaluate the structure and priors of a particular problem with various differentiable programs (models) using the characteristics defined in the previous sections. This problem is the classification of scientific publications into one of six classes with the CiteSeer dataset \cite{yang2016revisiting}. We use the PyTorch framework, especially the PyTorch Geometric \cite{PyGeometric2019} and self-attention library.

The CiteSeer dataset is a citation network extracted from the CiteSeer digital library. Nodes are publications and the edges denote citations between publications. The network has 3327 nodes, 3703 features and 6 classes for each node, 9104 edges and is undirected (the edges appearing twice in the edge matrix). The average degree is 2.737, the average clustering coefficient is 0.141 and there are isolated nodes. The dataset is divided into 120 training nodes, 500 validation nodes and 1000 test nodes.

\subsection{Results}
\label{sec52}
First, we use non-linear transformations (neural networks) of the vectors of each node without taking into account the graph structure (links):
\begin{enumerate}
\item Using a neural network (without any relational information) with two layers (input and output layer), 59,366 parameters, with node features as inputs and training with the training mask (120 nodes), the test accuracy is 58.2\%. The model suffers overfitting due to only a small amount of training nodes and because it does not take into account the link information. Therefore, it generalizes poorly to unseen node representations.
\item Using a neural network (without any relational information) with three layers and with node features as inputs, intermediate dimensions 512 and 256 (2,029,318 of parameters) and training with the training mask,
the test accuracy is 59\%, that is, slightly above the 58.2\% of the previous configuration. Using the validation mask (500 nodes) to train the model, the loss decreases more slowly and the test accuracy is around 68\%.
\end{enumerate}

Second, we apply graph neural networks by adding information from the neighboring nodes:
\begin{enumerate} 
\item Using a GNN with two graph convolutional layers and 59,366 parameters, the test accuracy is around 71.4\%. If we use the validation mask to train, the test accuracy is 76.2\%.
\item Using a GNN with three graph convolutional layers, we see that the test accuracy decreases to 62\% as the intermediate dimensions increase, probably because the model is overfitting the training data. If we lower the feature dimensions to 16, then the test accuracy returns to 67.4\%. If we use the validation mask to train the model, we get a test accuracy of 73.3\% with feature dimensions of 256 and 1,015,558 parameters.
\end{enumerate}

Third, we use a GNN with two graph convolutional layers and 59,366 parameters but modifying the number of training nodes to assess the influence of the training size:
\begin{enumerate}
\item When the first 620 nodes are used for training the GCN to avoid overfitting, we get a test accuracy of 77\%. We obtain the following training set-test accuracy: 50 nodes-62.1\%, 70 nodes-63.6\%, 120 nodes-71.4\%, 400 nodes-77\%, 500 nodes-77.3\%, 620 nodes-77\%, 800 nodes-76.80\%, 1000 nodes-76.2\%, 1500 nodes-76.9\%. Using the validation mask to train, we get a test accuracy of 76.2\%.
\end{enumerate}

Fourth, we use a GNN with two graph convolutional layers and 59,366 parameters but modifying the links (edges) of the original graph:
\begin{enumerate}
\item Using random edges (directed edges) between the nodes and training with the training mask, the test accuracy is 16.5\% compared to the 71.4\% obtained with the correct edges.
\item Using the training mask and a (i) 20\% of the edges removed gives a test accuracy 67.7\%; (ii) 25\% removed, 68.7\%; (iii) 33\% removed, 67.1\%; (iv) 50\% removed, 68.1\%; (v) 66\% removed, 62\%; (vi) 75\% removed, 61.1\%; and (vii) 80\% removed, 59.2\%; compared to the 71.4\% test accuracy obtained with the correct edges and the 58.2\% of the neural network.
\end{enumerate}

Finally, we design a model with a self-attention module between graph nodes, a pre-linear and post-linear layer for adjusting dimensions and a dropout layer. The model has 503,286 parameters and we modify the attention mask to prevent attention to certain positions:
\begin{enumerate}
\item We apply attention to all the nodes of the graph to learn the importance weights between the nodes. Then, training with the training mask the accuracy is 18.10\%, while training with the first 1500 nodes the accuracy is 23.10\%. 
\item We apply attention only between the same node. Then, training with the training mask the accuracy is 51.30\%, while training with the first 1500 nodes the accuracy is 69.10\%.
\item We apply attention only between neighboring nodes, thus respecting the graph structure and learning the weights between neighboring nodes. Then, training with the training mask the accuracy is 65.40\%, while training with the first 1500 nodes the accuracy is 73\%. 

\end{enumerate}

\subsection{Discussion}
\label{sec53}

The flexibility of differentiable programming, as a generalization of deep learning, allows us to use very diverse structures to model a specific problem. In Section \ref{sec42} We have evaluated the performance of the model with the  CiteSeer dataset by incorporating the structure of the problem to a different degree.

When we used conventional neural networks, without incorporating the links of the graph, the performance was acceptable (59\%). Increasing the number of layers and parameters of the neural network did not significantly increase accuracy.

When we applied GNNs by adding information from the neighboring nodes, the accuracy increased significantly, even more when we increased the number of training nodes (77\%). 

However, when we added more convolutional layers to the GNN, the test accuracy decreased. As we have seen in Section \ref{sec43}, in the graph-like CiteSeer dataset, with an average degree of 2.74, increasing the number of layers helps to propagate the information between distant nodes. But probably the task that we performed at each node or document (classification) only depends on the local neighborhood and not on distant nodes, so, in such cases, we do not need any more layers in the GNN.

Also, when we purposely changed the links of the graph when implementing the GNN, the accuracy decreased. When we used random links between the nodes, the accuracy was very bad (16.5\%). Removing an increasing percentage of the links lowered the accuracy, approaching the accuracy achieved using neural networks instead of GNNs.

Finally we evaluated the more general transformation, self-attention. As stated in \ref{sec41}, it transforms a set of vectors (nodes), dynamically choosing the most important ones.  The model has to learn the graph structure. When we applied attention to all the nodes of the graph to learn the importance weights between the nodes, the accuracy was very low (18.1\%). When we applied attention only between neighboring nodes, thus respecting the graph structure, accuracy went up to 73\%. Therefore, general differentiable programs such as self-attention, applied to problems with structure and without a great amount of data do not work well.

\section{The limits of differentiable models and the path to new learning strategies}
\label{sec6}

As we have seen, in differentiable models, the learning algorithm takes as input a sequence of training samples and outputs a function $g:\mathcal{X} \rightarrow \mathcal{Y} \in \mathcal{H}$, from the hypothesis space of functions, with a low probability of error $P(g(X)\neq Y$. These models are composed of any transformation of tensors and if the transformation has parameters to be learnt, it has to be differentiable. Then, differentiable models learn the transformations of the input that minimize the error to predict the training data. 

But the issue with such models is that they do not learn certain aspects of the task that are later combined to generate new knowledge. In particular, they cannot generate new tasks or information without additional training data. As a result, it is not possible to build an increasingly complex world model with just a differentiable model.

By contrast, humans can correctly interpret novel combinations of existing concepts, even if those combinations have not been seen previously. It could be said that differentiable models scale well in data (when we increase the input data) but not in tasks, while in humans it is the other way around.

We need learning strategies that leverage previous data and models to generate new experiences and knowledge. These new learning strategies should have the following characteristics.
\begin{enumerate} 
\item Be similar to the learning strategies developed by humans but adapted to the machine learning framework. As pointed out in \cite{Laird_Learning_2018}, humans have two levels of learning. Level 1, which is more automatic, continuous and unconscious, and level 2, which comprises the deliberate learning strategies that create the experiences for level 1.

\item In the framework of machine learning, level 1 corresponds to differentiable models that approximate a function and level 2 to learning strategies that, based on previous data and models, create new experiences and tasks. 

\item These learning strategies are algorithmic transformations of existing models and data. They build an increasingly complex model of the world and can be more logical (some logical combination of models characteristics or outputs) or automatic.

\item Given the training data $(x_i,x_i)$ and task $T_j$ for each model, a learning strategy is an algorithm $A:H_u \rightarrow g_u$, where $H_u$ are transformations of all previous information (training data, model information, etc.) and $g_u$ is a new model that, when presented with an input and a task not seen in the training data, is capable of getting the correct output for this task.

\item They can be based on any previous information: Priors of the models, information of the trained models (intermediate variables, output), training data, automatically generated labels, etc. 

\end{enumerate}

We describe here some illustrative examples of these strategies.
\begin{enumerate} 
\item Transformations of training data that scale in tasks. A large amount of data (e.g., text) is collected in the form of $(Task \, 1, data)...(Task \, n,data)$ and the differentiable model learns to map a bigger space: $probability(output | task; input; previous \, output)$. The model predicts an output by also conditioning on the task in addition to the input and previous output. In this way, a large amount of data is transformed and adapted to build differentiable models that scale in tasks.
GPT-X models \cite{radford2019language,brown2020language}, based on language models which are trained in an unsupervised manner on webpage datasets, are a great advance in this line.

\item Transformation using model priors. 
We first train one or several differentiable models (e.g., GNNs or attention models) used for some tasks. Each of these models has a series of internal components or variables with a physical meaning, $v^i_1,v^i_2,...,v^i_n$ for model $i$. Then we apply some machine learning technique (e.g., another differentiable model, symbolic regression, etc.) to extract some relations $R(v^1_1,...,v^i_n)$ between the variables of the models. In this way we can generalize and perform new tasks. For example, if we have several attentional convolutional models that perform different tasks on an image, we can relate the attention weights of the different models.

\item Control and guidance of a program. In an algorithm with a sequence of steps, some steps can depend on some pre-trained functions. Such is the case when decisions are guided by pre-trained models without reward. Algorithm 1 below illustrates this situation with the purchase or sell of raw materials based on the prediction of oil prices and interest rates.

\begin{algorithm}
  \caption{An algorithm for guided decisions based on pre-trained models}
\begin{algorithmic}
\setstretch{1.4}

\State $input 1 \gets oil(t-1,t)$ \algorithmiccomment{Get the oil prices input data}
\State $input 2 \gets rates(t-1,t)$
\algorithmiccomment{Get the interest rates input data}
\State $p1 \gets model1(input1)$ \algorithmiccomment{
Use a pre-trained model to predict the trend of oil prices}
\State $p2 \gets model2(input2)$ \algorithmiccomment{
Use a pre-trained model to predict the trend of interest rates}
\If{$(p1 > 0) \land (p2 > 0)$}
    purchase \algorithmiccomment{Buy if there is a positive trend in both assets}
\Else
\If{$(p1 \leq 0) \oplus (p2 \leq 0)$}
    don't do anything
\Else 
    \If{$(p1 \leq 0) \land (p2 \leq 0)$}
    sell
\EndIf
\EndIf
\EndIf

\end{algorithmic}
\end{algorithm}
\end{enumerate}

In sum, artificial intelligence models need to build an increasingly complex model of the world. In this section we have discussed the need, characteristics and examples of these new learning strategies. Nevertheless, much work remains to be done to characterize and specify the learning strategies that leverage existing knowledge (models and data), in a similar way as humans do. Furthermore, these strategies should be as automatic as possible and rely less on human engineering.

\section{Conclusions}
\label{sec7}
Differentiable programs are very flexible and give rise to a wide variety of structures. This flexibility makes it easy to generate the restricted hypothesis space, translating the prior knowledge and structure of the problem to the elements of a differentiable program, i.e., a tuple $\left \langle  n,l,E,f_i,v_i,\alpha_i \right \rangle$.

Similar to classical programming, it is useful
to define some characteristics to incorporate the knowledge and structure of the problem to be solved. These characteristics allow us to analyse the differentiable structures that are used in several domains or models and to define new structures with new capabilities. 

In experiments, we have seen that in problems with few training data it is important to incorporate the structure of the problem into the models rather than using very general models.

We have also seen that a shortcoming with differentiable models is that they just minimize an error to predict the data, but they cannot generate new tasks or information without additional training data. We have presented the characteristics and examples of new learning strategies that leverage existing knowledge (models and data) to generate new knowledge, in a similar way as humans. However much work remains to be done in this regard.

%%%%%%%%%%%%%%%%%%%%%%%%%%%%%%%%%%%%%%%%%%
\vspace{6pt} 

%%%%%%%%%%%%%%%%%%%%%%%%%%%%%%%%%%%%%%%%%%
%% optional
%\supplementary{The following are available online at \linksupplementary{s1}, Figure S1: title, Table S1: title, Video S1: title.}

%%%%%%%%%%%%%%%%%%%%%%%%%%%%%%%%%%%%%%%%%%
\authorcontributions{Writing--original draft preparation and experiments, A.H.; Writing--review, G.M. and J.M.A.; Content discussion and final version, A.H.,G.M. and J.M.A.}

%%%%%%%%%%%%%%%%%%%%%%%%%%%%%%%%%%%%%%%%%%
\funding{This work was supported by the Spanish Ministry of Science and Innovation, grant PID2019-108654GB-I00.}

%%%%%%%%%%%%%%%%%%%%%%%%%%%%%%%%%%%%%%%%%%
\acknowledgments{In this section you can acknowledge any support given which is not covered by the author contribution or funding sections. This may include administrative and technical support, or donations in kind (e.g., materials used for experiments).}

%%%%%%%%%%%%%%%%%%%%%%%%%%%%%%%%%%%%%%%%%%
\conflictsofinterest{The authors declare no conflict of interest. The funders had no role in the writing of the manuscript, or in the decision to publish the results.} 

{\small \textbf{Data Availability Statement:} The original data used for the simulations can be obtained from \url {https://github.com/kimiyoung/planetoid/tree/master/data}}

{\small \textbf{Computer Code and Software:} The developed source code is available at \url {https://github.com/adrianhernr/DPPaperExperiments}}

%%%%%%%%%%%%%%%%%%%%%%%%%%%%%%%%%%%%%%%%%%
%% optional
\abbreviations{The following abbreviations are used in this manuscript:\\

\noindent 
\begin{tabular}{@{}ll}
CNN & Convolutional neural network\\
FNN & Feed-forward neural network\\
GNN & Graph neural network\\
GPT-3 & Generative Pre-Trained Transformer-3\\
GPU & Graphics Processing Units\\
LSTM & Long short-term memory\\
RNN & Recurrent Neural Network
\end{tabular}}

%%%%%%%%%%%%%%%%%%%%%%%%%%%%%%%%%%%%%%%%%%
%% optional appendix


%=====================================
% References, variant A: internal bibliography
%=====================================
\reftitle{References}

%=====================================
% References, variant B: external bibliography
%=====================================
\externalbibliography{yes}
\bibliography{References}

%%%%%%%%%%%%%%%%%%%%%%%%%%%%%%%%%%%%%%%%%%
%% optional sample availability

%%%%%%%%%%%%%%%%%%%%%%%%%%%%%%%%%%%%%%%%%%
\end{document}